\documentclass[10pt,twocolumn,letterpaper]{article}

\usepackage{iccv}
\usepackage{times}
\usepackage{epsfig}
\usepackage{graphicx}
\usepackage{amsmath}
\usepackage{amssymb}
\usepackage{pifont}
\usepackage{hhline}
\usepackage{multirow}

\usepackage[pagebackref=true,breaklinks=true,letterpaper=true,colorlinks,bookmarks=false]{hyperref}

 \iccvfinalcopy 

\def\iccvPaperID{1972} 

\newcommand{\videoNum}{2,883}
\newcommand{\insNum}{4,883}
\newcommand{\trainNum}{2,238}
\newcommand{\valNum}{302}
\newcommand{\testNum}{343}
\ificcvfinal\fi
\newcommand{\specialcell}[2][c]{%
  \begin{tabular}[#1]{@{}c@{}}#2\end{tabular}}
  
\newcommand{\cmark}{\ding{51}}%
\newcommand{\xmark}{\ding{55}}%
\begin{document}

\title{Video Instance Segmentation}

\author{Linjie Yang\thanks{This work is partially done when Linjie is with Snap Inc.}\\
ByteDance AI Lab\\
{\tt\small linjie.yang@bytedance.com}
\and
Yuchen Fan\\
UIUC\\
{\tt\small yuchenf4@illinois.edu}
\and
Ning Xu\\
Adobe Research \\
{\tt\small nxu@adobe.com}
}

\maketitle

\begin{abstract}
   In this paper we present a new computer vision task, named video instance segmentation. The goal of this new task is simultaneous detection, segmentation and tracking of instances in videos. In words, it is the first time that the image instance segmentation problem is extended to the video domain. To facilitate research on this new task, we propose a large-scale benchmark called YouTube-VIS, which consists of {\videoNum} high-resolution YouTube videos, a 40-category label set and 131k high-quality instance masks. In addition, we propose a novel algorithm called MaskTrack R-CNN for this task. Our new method introduces a new tracking branch to Mask R-CNN to jointly perform the detection, segmentation and tracking tasks simultaneously. Finally, we evaluate the proposed method and several strong baselines on our new dataset. Experimental results clearly demonstrate the advantages of the proposed algorithm and reveal insight for future improvement. We believe the video instance segmentation task will motivate the community along the line of research for video understanding. 
\end{abstract}


\section{Introduction}

\begin{figure}[t]\centering
\includegraphics[width=1.0\linewidth]{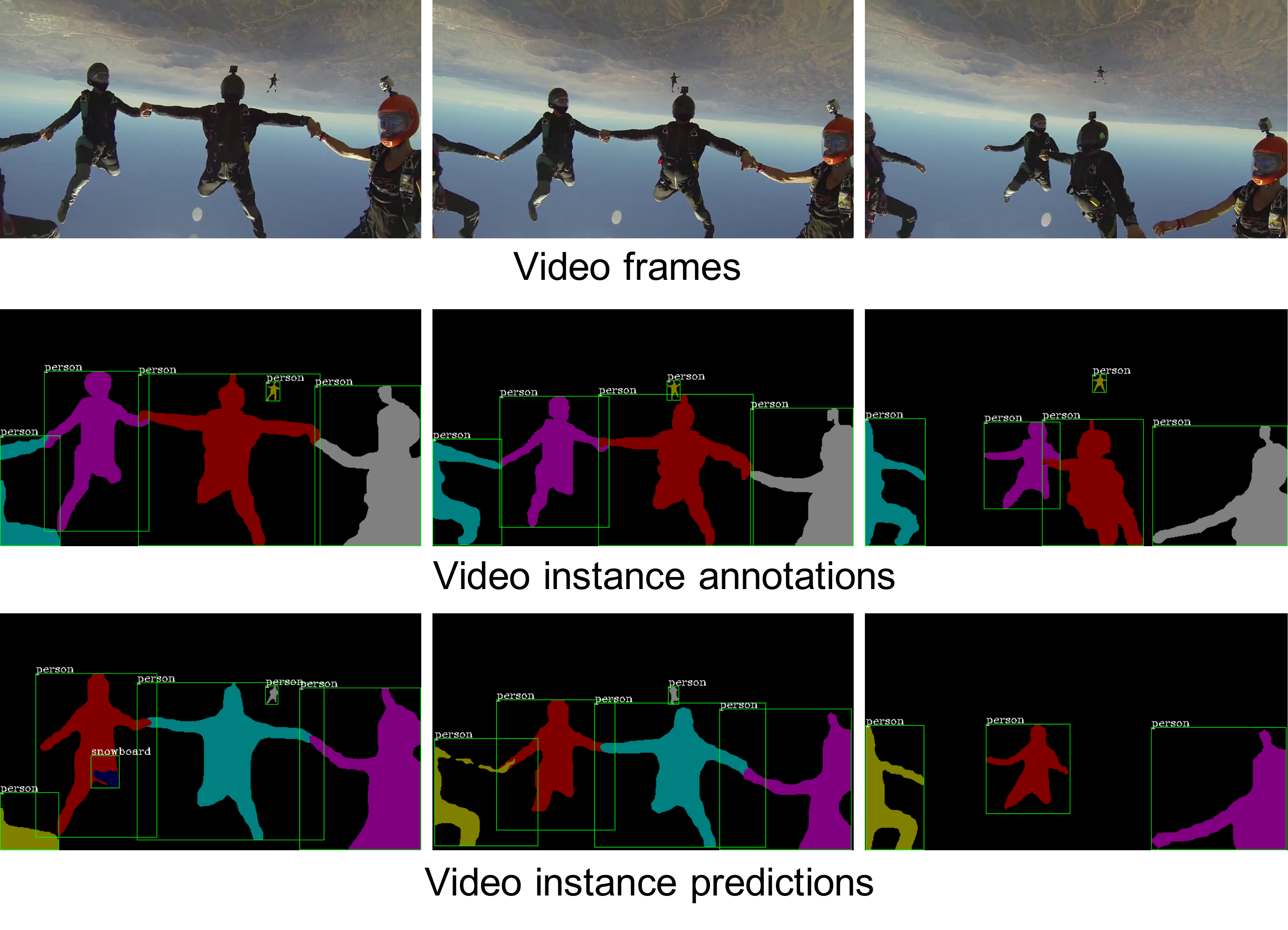}
\caption{A illustration of video instance segmentation. The three rows show image frames in a video, video instance annotations, and video instance predictions by our algorithm respectively. Masks in same color belong to the same object instance. Ground truth and predicted object categories are given on top of each bounding box.}
\label{fig:overview}\vspace{-2pt}
\end{figure}

Segmentation in images and videos is one of the fundamental problems in computer vision. In the image domain, the task of instance segmentation, \ie simultaneous detection and segmentation of object instances in images, was first proposed by Hariharan~\etal~\cite{hariharan2014simultaneous} and since then has attracted tremendous amount of attention in computer vision due to its importance. In this paper, we extend the instance segmentation problem in the image domain to the video domain. Different from image instance segmentation, the new problem aims at \textbf{simultaneous detection, segmentation and tracking of object instances in videos}. Figure~\ref{fig:overview} illustrates a sample video with ground truth annotations for this problem. Naturally, we name the new task \textbf{video instance segmentation}. The new task opens up possibilities for applications which requires video-level object masks such as video editing, autonomous driving and augmented reality. To our best knowledge, this is the first work to address video instance segmentation problem.
\vspace{-2pt}

Video instance segmentation is more challenging than image instance segmentation in that it not only requires instance segmentation on individual frames, but also the tracking of instances across frames. On the other hand, video content contains richer information than a single image such as motion pattern of different objects and temporal consistency, and thus provides more cues for object recognition and segmentation. Video instance segmentation is also related to several existing tasks. For example, video object segmentation~\cite{Caelles17osvos,Perazzi2016davis,Pont-Tuset2017davis} aims at segmenting and tracking objects in videos, but does not require recognition of object categories. Video object detection aims at detecting and tracking objects, but does not deal with object segmentation. 

One potential reason that video instance segmentation is seldomly studied is the lack of a large-scale dataset. Despite the existence of video segmentation datasets~\cite{Cordts2016Cityscapes,Pont-Tuset2017davis,xu2018ytvos} for other tasks, none of them is directly applicable to video instance segmentation. Given a video, our task requires both the masks of all instances of a predefined category set and the instance identities across frames to be labeled. Existing video segmentation datasets either do not have exhaustive labeling~\cite{Pont-Tuset2017davis,xu2018ytvos}, or do not have the object identities~\cite{Cordts2016Cityscapes}. Therefore, in this paper, we present the first large-scale dataset, named \textit{YouTube-VIS}, for video instance segmentation. The new dataset contains $\videoNum$ high-resolution YouTube videos, a 40-category label set including common objects such as person, animals and vehicles, $\insNum$ unique video instances and 131k high-quality masks. Our new dataset can be served as a benchmark for not only the video instance segmentation task, but also related tasks such as video semantic segmentation and video object detection.

In addition, we propose a novel algorithm called \textit{MaskTrack R-CNN} for video instance segmentation. Based upon Mask R-CNN~\cite{he2018maskrcnn} which is a state-of-the-art method for image instance segmentation, a new branch is added to the framework for tracking instances across video frames. Predicted instances are stored to an external memory and matched with objects in later frames. Moreover, we also propose several baselines by adapting top-performing methods from related tasks to our task, and compare their performance with our new method. Experimental results clearly demonstrate the advantage of our new algorithm and reveal insights for future improvement. Our dataset has been released at \url{https://youtube-vos.org/dataset/vis}. The code of our algorithm has been released at \url{https://github.com/youtubevos/MaskTrackRCNN}.

We conclude the contribution of this paper as follows.
\begin{itemize}
    \item To our best knowledge, it is the first time that video instance segmentation is formally defined and explored.
    \item We create the first large-scale video instance segmentation dataset which contains 2.9k videos and 40 object categories. 
    \item We propose a novel algorithm for video instance segmentation and compare it with several baselines on our new dataset.
\end{itemize}

The rest of our paper is organized as follows. In Section~\ref{sec:related} we briefly state the difference between related tasks and our new task. In Section~\ref{sec:task_definition} we formally introduce the video instance segmentation problem and evaluation metrics. Our new dataset and algorithm is elaborated in Section~\ref{sec:dataset} and~\ref{sec:alg} respectively. Finally, experimental results are presented in Section~\ref{sec:experiment}.

\section{Related Work}\label{sec:related}
Although video instance segmentation has been largely neglected in the literature, several related tasks have been well studied such as image instance segmentation, video object tracking, video object detection, video semantic segmentation and video object segmentation.

\noindent\textbf{Image Instance Segmentation} Instance segmentation not only group pixels into different semantic classes, but also group them into different object instances~\cite{hariharan2014simultaneous}. A two-stage paradigm is usually adopted, which first generate object proposals using a Region Proposal Network (RPN)~\cite{ren2015faster}, and then predict object bounding boxes and masks using aggregated RoI features\cite{dai2016mnc,Li2017fcis,he2018maskrcnn}. The proposed video instance segmentation not only requires segmenting object instances in each frame, but also determining the correspondence of objects across frames.

\noindent\textbf{Video Object Tracking} Video object tracking has two different settings. One is the detection-based tracking which simultaneously detect and track video objects. Methods~\cite{sadeghian2017tracking,wojke2017simple,son2017multi} under this setting usually take the ``tracking-by-detection" strategy.  The other setting is the detection-free tracking~\cite{bertinetto2016siamfc,nam2016mcnn,feichtenhofer2017det2track}, which targets at tracking objects given their initial bounding boxes in the first frame. Among the two settings, DBT is more similar to our problem as it also requires a detector. However, DBT only requires to produce bounding boxes, which is different from our task. Recently, a multi-object tracking and segmentation dataset~\cite{voi2019mots} is proposed to evaluate multi-object tracking along with instance segmentation. Their dataset is similar to our YouTube-VIS with respect to the exhaustive video instance annotations, but is greatly inferior to ours on the data scale and object categories.

\noindent\textbf{Video Object Detection} Video object detection aims at detecting objects in videos, which is first proposed as part of ImageNet visual challenge~\cite{ILSVRC15}. While object identity information are often utilized for improve robustness of detection algorithms~\cite{feichtenhofer2017det2track,zhu2017dff,zhu17fgfa}, the evaluation metric is limited to per-frame detection and does not require joint object detection and tracking. 

\noindent\textbf{Video Semantic Segmentation} Video semantic segmentation is a direct extension of semantic segmentation to videos, where image pixels are predicted as different semantic classes. Temporal information such as optical flow is adopted to improve either accuracy~\cite{zhu2017dff} or efficiency~\cite{zhu2017dff,li2018lvs,shelhamer2016clockwork} of semantic segmentation models. Video semantic segmentation does not require explicit matching of object instances across frames.

\noindent\textbf{Video Object Segmentation} Video object segmentation has gained substantial attention in recent years, which has two scenarios: semi-supervised and unsupervised. Semi-supervised video object segmentation~\cite{Perazzi2017masktrack,Caelles17osvos} targets at tracking and segment a given object with a mask. Visual similarity~\cite{Caelles17osvos,chen2018blazingly, voigtlaender2019feelvos}, motion cues~\cite{Cheng2017segflow}, and temporal consistency~\cite{Perazzi2017masktrack,Yang2018osmn} are extracted to identity the same object across the video. In unsupervised scenario,  a single foreground object is segmented~\cite{Tokmakov2016LearningMP,jain2017fusionseg,tokmakov2017lvo}.  In both settings, algorithms consider the target objects as general objects and does not care about the semantic categories.

\section{Video Instance Segmentation}\label{sec:task_definition}

\noindent\textbf{Problem Definition}. In video instance segmentation, we have a predefined category label set $\mathcal{C}=\{1,...,K\}$ where $K$ is the number of categories. Given a video sequence with $T$ frames, suppose there are $N$ objects belonging to the category set $\mathcal{C}$ in the video. For each object $i$, let $c^i\in\mathcal{C}$ denote its category label, and let $\mathbf{m}_{p...q}^i$ denote its binary segmentation masks across the video where $p\in[1,T]$ and $q\in[p,T]$ denote its starting and ending time. Suppose a video instance segmentation algorithm produces $H$ instance hypotheses. For each hypothese $j$, it needs to have a predicted category label $\tilde{c}^j\in\mathcal{C}$, a confidence score $s^j\in[0,1]$ and a sequence of predicted binary masks $\mathbf{\tilde{m}}_{\tilde{p}...\tilde{q}}^j$. The confidence score is used for our evaluation metrics which will be explained shortly. 

The goal of our task is minimizing the difference between the ground truth and the hypotheses. In other words, a good video instance segmentation method should be able to have a good detection rate of all instances, track all the instances reliably and localize the instance boundaries accurately. It should be noted that there is some minor difference between our task and the multi-object tracking problem~\cite{milan2016mot16} in that a still object instance is treated as a ground truth, and if an object is occluded or out of scene for several frames then reappears in the following frames, the instance label should be consistent.

\noindent\textbf{Evaluation Metrics}. We borrow the standard evaluation metrics in image instance segmentation with modification adapted to our new task. Specifically, the metrics are average precision (AP) and average recall (AR). AP is defined as the area under the precision-recall curve. The confidence score is used to plot the curve. AP is averaged over multiple intersection-over-union (IoU) thresholds. We follow the COCO evaluation to use 10 IoU thresholds from 50\% to 95\% at step 5\%. AR is defined as the maximum recall given some fixed number of segmented instances per video.
Both of the two metrics are first evaluated per category and then averaged over the category set. 

Our IoU computation is different from image instance segmentation because each instance contains a sequence of masks. To compute the IoU between a ground truth instance $\mathbf{m}_{p...q}^i$ and a hypothese instance  $\mathbf{\tilde{m}}_{\tilde{p}...\tilde{q}}^j$, we first extend $p$ and $\tilde{p}$ to 1, $q$ and $\tilde{q}$ to $T$ by padding empty masks. Then, 
\begin{align}
\text{IoU}(i,j) = \frac{\Sigma_{t=1}^T|\mathbf{m}_t^i\cap\mathbf{\tilde{m}}_t^j|}{\Sigma_{t=1}^T{|\mathbf{m}_t^i\cup\mathbf{\tilde{m}}_t^j|}}
\end{align}

The proposed IoU computes the spatial-temporal consistency of predicted and ground truth segmentations. If the algorithm detects object masks successfully, but fails to track the objects across frames, it will get a low IoU.

\section{YouTube-VIS}\label{sec:dataset}

Since none of the existing video segmentation datasets matches the requirement for our video instance segmentation task, we need to collect a new benchmark dataset for the development and evaluation of the proposed methods. There are several criteria that the new benchmark need to satisfy. First, it should contain common instance categories, just like recent image instance segmentation benchmarks~\cite{lin2014mscoco,hariharan2014simultaneous}. Second, it should contain video instances with various challenging cases such as occlusion, appearance change, heavy camera motion etc. Last but not the least, the annotation quality should also be high, which is a common problem in some of the existing segmentation datasets with polygon-based annotations.

\begin{figure}[t]\centering
\includegraphics[width=1\linewidth]{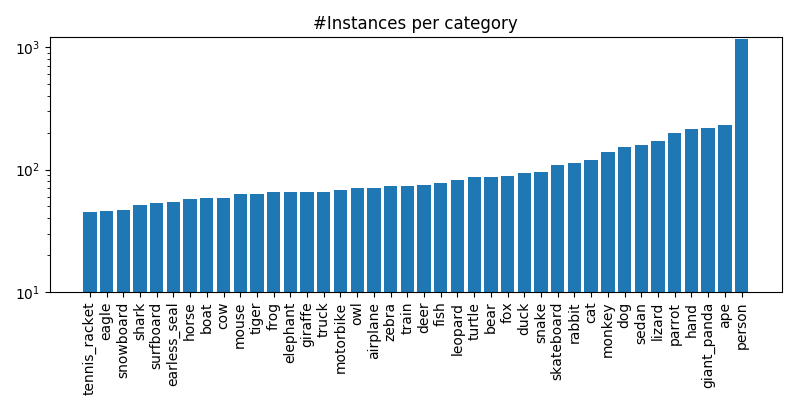}
\caption{Number of unique video objects for the 40 categories in our dataset. }
\label{fig:objects}
\end{figure}

With the above criteria in mind, we create a new large-scale benchmark called \textit{YouTube-VIS}. Instead of building our benchmark from scratch, we take advantage of an existing dataset called YouTube-VOS~\cite{xu2018ytvos}. YouTube-VOS is a large-scale video object segmentation dataset which is comprised of 4453 high-resolution YouTube videos and 94 common object categories. In each video, several objects are labeled by tracing the object boundaries manually at every 5 frames in a 30fps frame rate. The length of each video is around 3 to 6 seconds. Even though object masks are not exhaustively labeled in YouTube-VOS, it still serves as a very good resource to build our own dataset. Specifically, we first select 40 common category labels from the 94 category labels as our category set. Then we sample around 2.9k videos with objects from the 40 categories in YouTube-VOS. We then ask human annotators to carefully label the other objects belonging to the category set exhaustively in these videos. As a result, our dataset is annotated with $\insNum$ unique objects and approximately 131k object masks. A comparison of some high level statistics of YouTube-VIS and related datasets are shown in Table~\ref{tab:stats}. The distribution of the unique objects per category in our dataset is illustrated in Figure~\ref{fig:objects}.

Our new dataset \textit{YouTube-VIS} is not only the first large-scale benchmark for video instance segmentation, but also a useful benchmark for other vision tasks such as video object detection and video semantic segmentation. It also complements the original YouTubeVOS dataset with more objects. We believe our new dataset will serve as a useful benchmark for various pixel-level video understanding tasks.

\begin{table}[t]
\centering
\footnotesize
\caption{High level statistics of YouTubeVIS and previous video object segmentation datasets. YTO, YTVOS, and YTVIS stands for YouTubeObjects,  YouTubeVOS, and YouTube-VIS respectively.}
\label{tab:stats}
\begin{tabular}{|l|l|l|l|l|l|l|l|}
\hline
              & \specialcell{YTO \\ \cite{Jain2014youtube}   }                & \specialcell{ FBMS\\ \cite{ochs2013fbms} }                 & \multicolumn{2}{l|}{\specialcell{ DAVIS\\ \cite{Perazzi2016davis, Pont-Tuset2017davis} }     }                       & \specialcell{YTVOS\\ \cite{xu2018ytvos} }    & YTVIS                \\ \hline
Videos        & 96                    & 59                    & 50                    & 90                                  & 4,453                  & \videoNum                  \\ \hline
Categories    & 10                    & 16                    & -                     & -                                  & 94                    & 40                    \\ \hline
Objects & 96                    & 139                   & 50                    & 205                                  & 7,755                  & \insNum                 \\ \hline
Masks         & 1.7k                  & 1.5k                  & 3.4k                 & 13.5k                               & 197k                  & 131k                  \\ \hline
Exhaustive    & \xmark & \xmark & \xmark & \xmark  & \xmark & \cmark \\ \hline
\end{tabular}
\vspace{-5pt}
\end{table}

\section{MaskTrack R-CNN}\label{sec:alg}
\begin{figure*}[t]\centering
\includegraphics[width=0.9\linewidth]{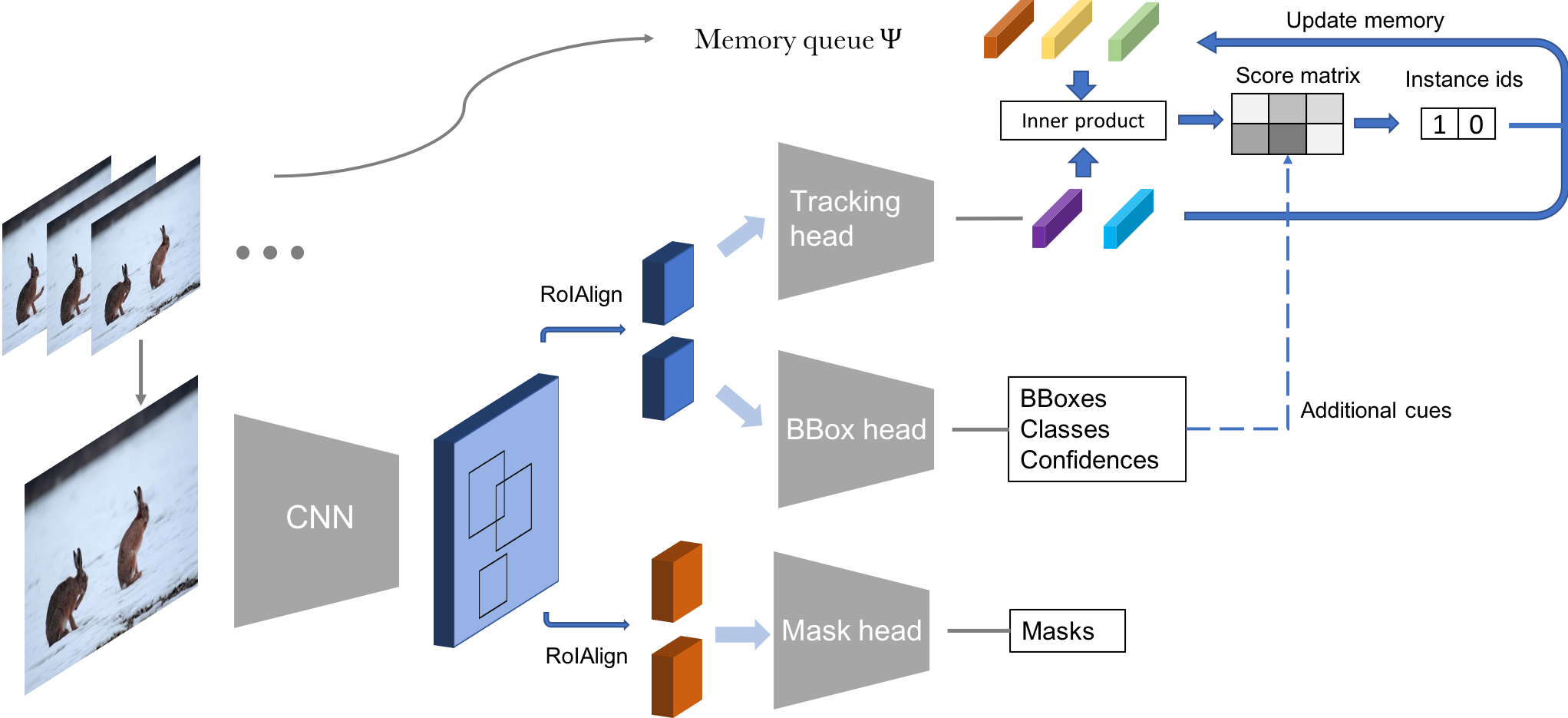}
\caption{An overview of our approach. A tracking head is embedded in the MaskRCNN framework to facilitate identity tracking of object instances through interaction with a memory queue. The memory queue is used to maintain all the existing object instances in the video. }
\label{fig:model}\vspace{-2pt}
\end{figure*}

Our new algorithm for video instance segmentation is built based on Mask R-CNN~\cite{he2018maskrcnn}. In addition to its original three branches for object classification, bounding box regression, and mask generation, we add the forth branch together with an external memory to track object instances across frames. The tracking branch mainly leverages the cue of appearance similarity. In addition, we propose a simple yet effective method to combine it with the other cues such as semantic consistency and spatial correlation to improve the tracking accuracy substantially. The overall framework of our algorithm is illustrated in Figure~\ref{fig:model}. For inference, our method processes video frames sequentially in an online fashion. Next we first briefly review Mask R-CNN and then describe our new components in detail.

\subsection{Mask R-CNN}
Mask R-CNN is a high-performing method for image instance segmentation. It consists of two stages. In the first stage, a RPN~\cite{ren2015faster} takes an image as input and proposes a set of candidate object bounding boxes. In the second stage, features are extracted by the RoIAlign operation from each candidate box and further used to perform classification, bounding box regression and binary segmentation in parallel by three dedicated branches. Please refer to~\cite{he2018maskrcnn} for more details.

\subsection{New Tracking Branch}

Our network adopts the same two-stage procedure, with an identical first stage which proposes a set of object bounding boxes at each frame. In the second stage, in parallel to the three branches (\ie classification, bounding box regression, binary segmentation), we add the forth branch to assign an instance label to each candidate box. Suppose there are already $N$ instances identified by our algorithm from previous frames. Then a new candidate box can only be assigned to one of the $N$ identities if it is one of the previous instances or a new identity if it is a new instance. We formulate the problem as multiclass classification. There are $N+1$ classification digits which represent the $N$ already identified instance and a new unseen instance which is denoted by digit 0. The probability of assigning label $n$ to a candidate box $i$ is defined as 
\begin{align}
    p_i(n) = 
    \begin{cases}
    \frac{e^{\mathbf{f}_i^\intercal\mathbf{f}_n}}{1+\Sigma_{j=1}^N{e^{\mathbf{f}_i^\intercal\mathbf{f}_j}}} & n \in [1,N] \\
    \frac{1}{1+\Sigma_{j=1}^N{e^{\mathbf{f}_i^\intercal\mathbf{f}_j}}} & n = 0
    \end{cases}
     \label{eqn:prob}
\end{align}
where $\mathbf{f}_i$ and $\mathbf{f}_j,j\in[1,N]$ denote the new features extracted by our tracking branch from the candidate box and the $N$ identified instances. Our tracking branch has two fully connected layers which project the feature maps extracted by RoIAlign into new features. Since the features of previously identified instances have already been computed, we use an external memory to store them for efficiency. The cross entropy loss is used for out tracking branch, \ie $L_{track}=-\Sigma_i\log(p_i(y_i))$ where $y_i$ is the ground truth instance label.

We dynamically update our external memory when a new candidate box is assigned with an instance label. If the candidate box belongs to an existing instance, we update the instance feature stored in the memory with the new candidate feature, which represent the latest state of the instance. If the candidate object is assigned with label 0, we insert feature of the candidate object to the memory and add 1 to the number of identified instances. 

We need a sequence of frames to train the new tracking branch. In our implementation we use a pair of frames which are randomly sampled from a training video. One of the frames is randomly picked as the reference frame while the other one is picked as the query frame. On the reference frame, we do not generate any candidate boxes but only extract features from its ground truth instance regions and save them into our external memory. On the query frame, candidate boxes are generated in the first stage and only positive candidate boxes are then matched to the instance labels in the memory and contribute to the tracking loss. A positive candidate box is the one with at least 70\% IoU overlapping with any ground truth object boxes. Our whole network is trained end-to-end with losses combined from the four branches together $L=L_{cls}+L_{box}+L_{mask}+L_{track}$.

\subsection{Combining Other Cues}
Our tracking branch computes the probability of assigning an instance label to a candidate box based on the appearance similarity. However, there are also other information such as semantic consistency, spatial correlation and detection confidence which could be leveraged to determine the instance labels. We propose a simple yet effective way to combine all these cues together to improve the tracking accuracy in a post-processing way. 

Specifically, for a new candidate box $i$, let $b_i$, $c_i$ and $s_i$ denote its bounding box prediction, category label and detection score, which are obtained from the bounding box branch and the classification branch of our network. Similarly, for an identified instance with label $n$, let $b_n$ and $c_n$ denote its bounding box prediction and category label associated with the saved features in the memory. Then a score for assigning the label $n$ to candidate box $i$ is computed as 
\begin{align}
    v_i(n) = \log p_i(n) + \alpha \log s_i +\beta 
    \text{IoU}(b_i, b_n) + \gamma \delta(c_i, c_n)
    \label{eq:score}
\end{align}
where $p_i(n)$ is obtained by Equation~\ref{eqn:prob}, $\text{IoU}(b_i, b_n)$ computes the IoU between $b_i$ and $b_n$, and $\delta(c_i, c_n)$ is a Kronecker delta function which equals 1 when $c_i$ and $c_n$ are equivalent and 0 otherwise. $\alpha$, $\beta$ and $\gamma$ are hyperparameters to balance the effect of different cues. Empirically we find that the score is not sensitive to different values of $\alpha$ and $\beta$.

Note that Equation~\ref{eq:score} is only used in the testing stage and does not contribute to the training of our network. There are also other possible ways to integrate these cues, for example, take all the cues as inputs and train an end-to-end network, which will be left as an interesting future study for us. 

\subsection{Inference}
Given a new testing video, our external memory is set empty and the number of identified instances is set 0. Our method processes each frame sequentially in an online fashion. At each frame, our network first generates a set of instance hypotheses. Non-Max Suppression (NMS) (50\% overlapping threshold) is applied to reduce the hypotheses. Then the remaining hypotheses are matched to identified instances from previous frames by Equation~\ref{eq:score}. Note that we do not match hypotheses within a single frame to avoid conflicts. All instance hypotheses of the first frame are directly regarded as new instances and saved into the external memory. It is possible for our method to match multiple hypotheses from a single frame to one instance label, which contradicts the common sense. We handle this case by keeping only one hypothese which has the largest score $v$ among the conflicting hypotheses while discarding the others.

After processing all frames, our method produces a set of instance hypotheses, each of which contains an unique instance label, and a sequence of binary segmentation, category labels and detection confidence. We use the averaged detection confidence as the confidence score for the whole sequence and use the majority votes of category labels as the final category label for the instance.

\section{Experiments}\label{sec:experiment}

In this section, we compare our MaskTrack R-CNN with several baselines on our new dataset YouTube-VIS. We first present the information of the dataset splits and implementation details of our method. 

\noindent\textbf{Dataset}. We randomly split the YouTube-VIS dataset into $\trainNum$ training videos, $\valNum$ validation videos and $\testNum$ test videos. Each of the validation and test set is guaranteed to have more than 4 instances per category. All the methods are trained on the training set and all hyperparameters are cross validated on the validation set. We present results on both the validation set and test set in the results section. 

\noindent\textbf{Implementation}. The backbone of our network is based on the network structure of ResNet-50-FPN in~\cite{he2018maskrcnn} and we use a public implementation~\cite{mmdetection2018} which is pretrained on MS COCO~\cite{lin2014mscoco}. The structure of our new tracking branch is two fully connected layers. The first fully connected layer transforms the $7\times7\times256$ input feature maps to 1-D $1024$ dimensions. The second fully connected layer also maps its input to 1-D $1024$ dimensions. Our full model is trained end-to-end in 12 epochs. The initial learning rate is set to 0.05 and decays with a factor of 10 at epoch 8 and 11. In testing, our model runs at 20 FPS with a NVIDIA 1080Ti GPU. The hyperparameters $\alpha$, $\beta$ and $\gamma$ in Equation~\ref{eq:score} are cross validated and chosen to be 1, 2 and 10 to produce our final results. We downsample original frame sizes to $640\times360$ for all the methods in both training and evaluation.

\subsection{Baselines} 
To our best knowledge, there is no prior work directly applicable to our new task. Therefore we combine ideas from related tasks to propose several new baselines. We incorporate two types of algorithms for the baselines. The first type uses the object masks detected in the first frame of the video as initial guidance and applies video object segmentation algorithms to propagate the masks. We evaluate two recent video object segmentation algorithms OSMN~\cite{Yang2018osmn} and FEELVOS~\cite{voigtlaender2019feelvos}. The second type follows the ``tracking-by-detection" idea which is very popular in the multi-object tracking task. The basic idea of this type of works is using image detection methods on each frame independently and then linking the detection across frames by various tracking methods. In our experiment, all the baselines are given the same per-frame instance segmentation results which are produced by a Mask R-CNN. The Mask R-CNN has the same structure as our network except the tracking branch. To make the evaluation fair, The Mask R-CNN is pretrained on MS COCO and is then finetuned on YouTube-VIS with 12 training epochs. Next we describe different track-by-detect methods in our experiment.

\noindent\textbf{IoUTracker+}. This method computes a score between a new candidate box with each identified instance by using a similar equation as Equation~\ref{eq:score} except without using the first term, \ie the appearance similarity. Therefore the matching does not leverage any visual information. The candidate box is assigned to the instance label with the largest score, with a minimum IoU threshold (30\%). Otherwise it is assigned with a new label. The matching process is similar to IoUTracker~\cite{bochinski2017high}. The difference is that a similar memory as our method is equipped with the baseline to save the information of identified instances. 

\noindent\textbf{OSMN~\cite{Yang2018osmn}}. Given an identified instance mask, OSMN estimates a new mask of the instance at a new frame. The new mask is then used to compute IoU with candidate boxes at the same frame. This is better than IoU directly computed via consecutive frames especially when an instance is occluded or has large motion. The rest of the matching process is the same as IoUTracker+.

\noindent\textbf{DeepSORT~\cite{wojke2017simple}}. DeepSORT is a top-performing tracking method. it uses Kalman filter to predict bounding box location to avoid directly computing IoU of consecutive frames. In addition, it use a deep network to measure the appearance similarity between bounding boxes. Finally the IoU score and the visual appearance score are combined to match tracks by the Hungarian algorithm.

\noindent\textbf{SeqTracker}. This is an offline algorithm following Seq-NMS~\cite{seqnms16}. Given a video and a set of instance segmentation results of every frame, SeqTracker searches all possible tracks to find the one with the largest score, which is computed similarly as IoUTracker+. Then the instance segmentation of the track will be removed from the set and the search process repeats. The method halts until the length of a retrieved track is less than a threshold, which is set to 8 in our experiment.

\begin{table*}[t]
\centering
\caption{Quantitative evaluation of the proposed algorithm and baselines on the YouTube-VIS validation and test set. The best results are highlighted in bold.}
\label{tab:compare}
\begin{tabular}{|l|c|c|c|c|c|c|c|c|c|c|c|}
\hline
\multicolumn{2}{|c|}{\multirow{2}{*}{\textbf{Methods}}}  & \multicolumn{5}{c|}{\begin{tabular}[c]{@{}c@{}} \textbf{validation set}\end{tabular}} &
\multicolumn{5}{c|}{\begin{tabular}[c]{@{}c@{}} \textbf{test set}\end{tabular}} \\ \cline{3-12}
\multicolumn{2}{|c|}{} & AP & AP\textsubscript{50} & AP\textsubscript{75} & AR\textsubscript{1} & AR\textsubscript{10}
 & AP & AP\textsubscript{50} & AP\textsubscript{75} & AR\textsubscript{1} & AR\textsubscript{10} \\ \hline
\multirow{2}{*}{\textbf{Mask propagation}}  & OSMN~\cite{Yang2018osmn} & 23.4 & 36.5 & 25.7 & 28.9 &31.1 & 27.3 & 44.4 & 28.0 & 28.8 & 34.0 \\ \cline{2-12}
 & FEELVOS~\cite{voigtlaender2019feelvos} & 26.9 & 42.0& 29.7& 29.9 & 33.4 & 29.6 & 45.4  & 30.7& 33.4 & 36.8 \\ \hline
\multirow{5}{*}{\textbf{Track-by-detect}} & IoUTracker+   & 23.6 & 39.2 & 25.5 & 26.2 & 30.9    & 25.2 & 41.9 & 26.2 &  28.7 & 33.7 \\ \cline{2-12}
& OSMN~\cite{Yang2018osmn}       & 27.5 & 45.1 & 29.1 & 28.6 & 33.1    & 27.3 & 44.4 & 28.0 & 28.8 & 34.0 \\ \cline{2-12}
& DeepSORT~\cite{wojke2017simple}   & 26.1 & 42.9 & 26.1 & 27.8 & 31.3    & 27.2 & 44.0 & 29.2 & 29.1 & 33.3 \\ \cline{2-12}
& SeqTracker    & 27.5 & 45.7 & 28.7 & 29.7 & 32.5    & 29.5 & 48.1 & 31.2 & 32.0 & 34.5 \\ \cline{2-12}
& MaskTrack R-CNN & \textbf{30.3} & \textbf{51.1} & \textbf{32.6} & \textbf{31.0} & \textbf{35.5}    & \textbf{32.3} & \textbf{53.6} & \textbf{34.2} & \textbf{33.6} & \textbf{37.3} \\ \hline
\end{tabular}
\end{table*}

\begin{figure*}[t]\centering
\includegraphics[width=0.95\linewidth]{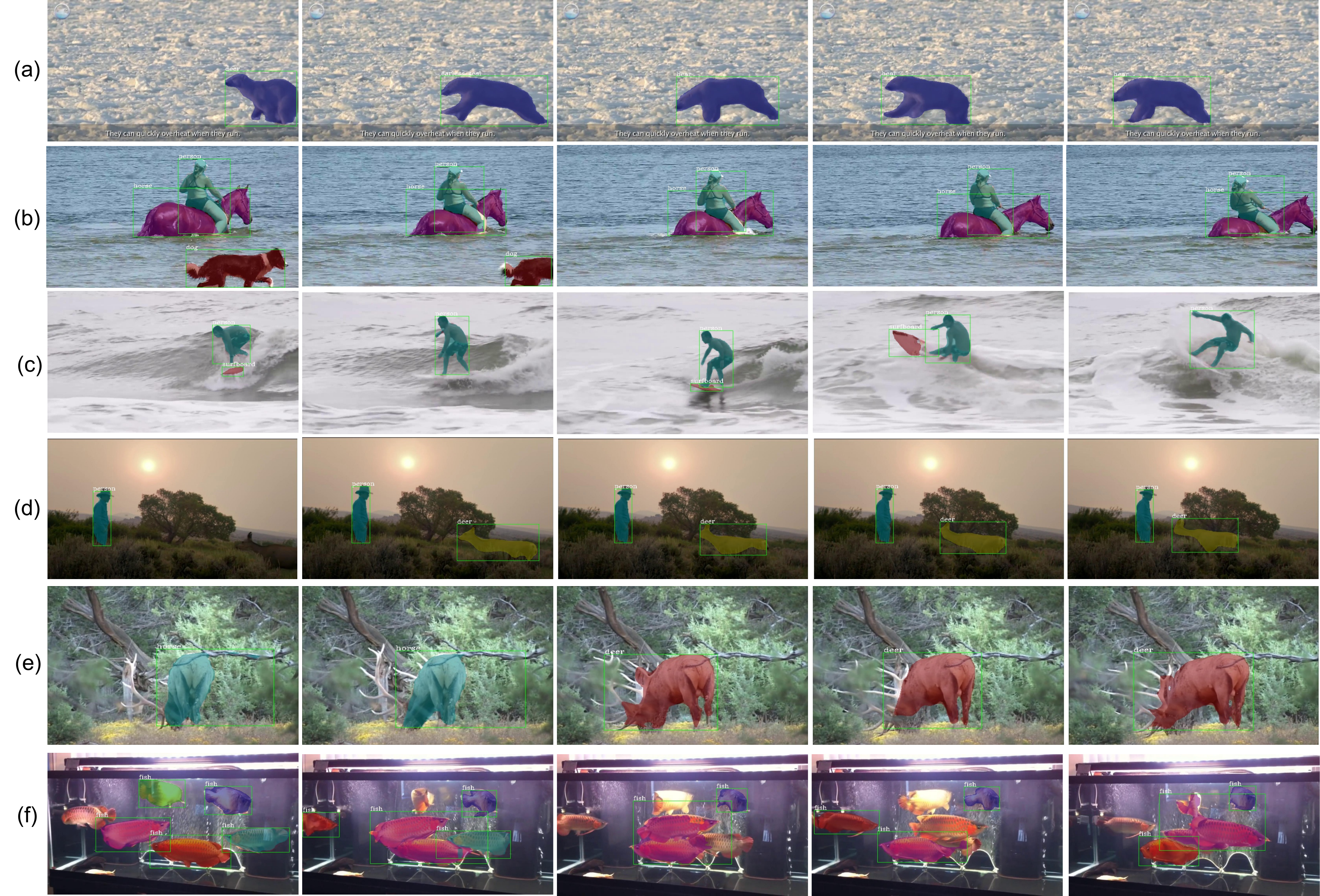}
\caption{Sample results of MaskTrack R-CNN. Each row have five sampled frames from a video sequence. (a),(b),(c) and (d) show correct predictions while (e) and (f) are failure cases. Objects with same predicated identity have the same color. Object category is shown on top of each bounding box. Zoom in to see details.}
\label{fig:sample}\vspace{-2pt}
\end{figure*}

\subsection{Main Results}
Table~\ref{tab:compare} presents the comparison results. Notably, our method MaskTrack R-CNN achieves the best results under all evaluation metrics and on both the validation and test sets. The main difference between our method and the other track-by-detect baselines is the new tracking branch which is trained end-to-end with the other branches, so that useful information can be shared among multiple tasks. The key for the joint training of tracking with other tasks is that we formulate the instance matching process as a differentiable component, which enables the matching loss to be properly back propagated. 

Next we analyze the performance of the baselines. For mask propagation algorithms, they suffer from a natural disadvantage, that they cannot handle objects appear in the intermediate frames. Also the flawed detections in the frist frames directly degenerate their performance. Even the state-of-the-art video object segmentation algorithm FEELVOS only gains 26.9 AP on validation set. For track-by-detect algorithm, IoUTracker+ does not leverage any visual information which is not surprising to gain weak performance. OSMN predicts the possible location of previously identified instances at new frames, and use the prediction to match instances, which is useful to handle occlusion and fast motion. DeepSORT improves IoUTracker+ on both the IoU matching and usage of visual similarity, achieving better results. SeqTracker does not depend on any visual information and achieves better performance than the other baselines. However, it is an offline method which requires instance segmentation results to be precomputed for all frames. The other methods including MaskTrack R-CNN are online methods, which produce instance tracks sequentially. 


Figure~\ref{fig:sample} shows six sample videos with our predictions. The first four rows ((a),(b), (c) and (d)) are successful predictions and the last two rows are failure cases. In video (a), the frame-level prediction gives incorrect results in the first two frames, where the bear is predicted as ``deer'' and ``earless seal''. The video-level prediction corrects these mistakes by majority voting of all frames. In video (c), the surfboard is occluded by the wave in multiple frames, our algorithm is able to track the surfboard after it disappears and reoccurs. The memory queue in MaskTrack R-CNN is able to keep track of all previous objects even they are disappeared in intermediate frames. In video (d), we show a case that new object enters the video in the intermediate frames. Our algorithm is able to detect the deer in the second frame as new object and add it to the external memory.
Video (e) and (f) shows two challenging cases. In video (e), the dear has quite different appearance in different poses, and our algorithm fail to recognize the same object and consider them as two different objects. In video (f), multiple similar fishes move around the aquarium and occlude each other. Our algorithm groups two fishes as one in the second and third frame, and gets confused with the object identities later on. 

\begin{figure}[t]\centering
\includegraphics[width=1\linewidth]{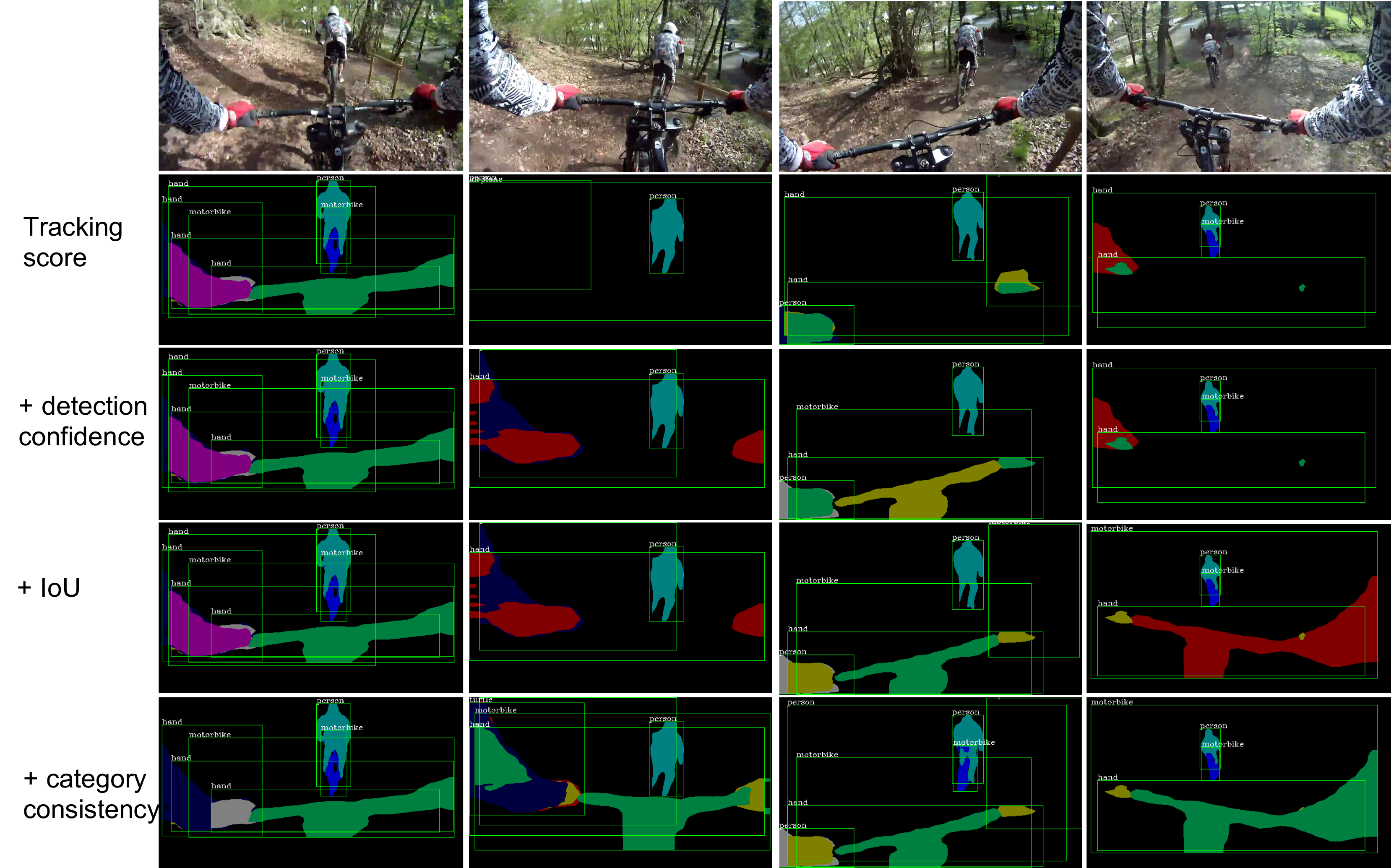}
\caption{A sample result with different matching cues used. With all four factors, the result is the best.}
\label{fig:factor}\vspace{-2pt}
\end{figure}

\subsection{Ablation Study}

\begin{table}[t]
\small
\centering
\caption{Ablation study of our method on the YouTube-VIS validation set. ``Det", ``IoU", and ``Cat" denote the detection confidence, the bounding box IoU, and the category consistency in Equation~\ref{eq:score} respectively. Numbers in brackets shows the difference compared to the complete score.}
\label{tab:combination}
\begin{tabular}{|l|l|l|l|l|l|}
\hline
Det & IoU & Cat & AP      & AP\textsubscript{50} & AP\textsubscript{75} \\ \hline
\xmark         & \xmark        & \xmark         & 21.1(-9.2)   & 37.7(-13.4)   & 23.6(-9.0)\\ \hline
\cmark         & \xmark        & \xmark         & 23.4(-6.9)   & 42.5(-8.6)   & 24.4(-8.2)\\ \hline
\xmark         & \cmark        & \xmark         & 22.7 (-7.6)   & 40.7 (-10.4)   & 25.2 (-7.4)\\ \hline
\cmark         & \cmark        & \xmark         & 24.7 (-5.6)   & 44.3 (-6.8)   & 26.7 (-5.9)\\ \hline
\xmark         & \xmark        & \cmark         & 27.9 (-2.4)   & 47.1 (-4.0)   & 30.5 (-2.1)\\ \hline
\cmark         & \xmark        & \cmark         & 29.2 (-1.1)   & 49.2 (-1.9)   & 31.9 (-0.7)\\ \hline
\xmark         & \cmark        & \cmark         & 29.5 (-0.8)   & 48.7 (-2.4)   & 32.2 (-0.4) \\ \hline
\cmark         & \cmark        & \cmark         & 30.3          & 51.1          & 32.6 \\ \hline
\end{tabular}
\end{table}

We study the importance of three cues used in Equation~\ref{eq:score} to our method. They are the detection score, the bouding box IoU and category consistency. We evaluate our method on the validation set by turning these cues on and off. The results are presented in Table~\ref{tab:combination}. We find that the bounding box IoU and the category consistency are most important to the performance of our method. Without any of them, AP will drop around 5\%. While the detection confidence score only improves our method slightly. Intuitively, the bounding box IoU correlates to the spatial relationship between instances, which is a strong prior in many cases. The category consistency also provides a very strong constraint because the category label of an instance should not change in a video. However, relying too much on the these factors can also cause problems due to the imperfect estimation. Therefore our method uses these cues as soft constraints. To visualize the effect of these three factors, we also generate predictions with the three factors added one by one on one specific sample, which is shown in Figure~\ref{fig:factor}. Note that the first three variants cannot track the identity of the ``green'' motorbike well, while the variant with four different cues is able to track it through the whole video.


\subsection{Oracle Results}

Additionally, we investigate the effectiveness of the two parts in our algorithm: image-level prediction and cross-frame association. We evaluate effectiveness of video-level association by applying ground truth image-level annotations to our algorithm. Specifically, given ground truth image-level predictions including bounding boxes, masks and categories, we compute matching score $p_i$ using RoIAlign features of ground truth bounding boxes and match objects across frames using combined score $v_i$. The result is shown in Table~\ref{tab:oracle} with ``Image Oracle''. We also evaluate image-level predictions with ground truth object identities. Towards this end, per-frame predictions are first matched to their closest ground truth image objects, and then video objects are aggregated using ground truth object identities. The result is shown in Table~\ref{tab:oracle} with ``Identity Oracle''. It shows that Image Oracle achieves much better performance than Identity Oracle, which means image-level predictions is critical for better performance on video instance segmentation. Identity Oracle is only marginally better than MaskTrack RCNN, which indicates limited potential of improving over our current method by modifying object tracking method. Improving image-level detection performance by utilizing properly-designed spatial-temporal feature could be a promising direction. Meanwhile, even with image-level ground truth, it is still challenging to associate objects across frames due to object occlusions and fast motion.

\begin{table}[t]
\centering
\caption{Oracle results in two settings on validation set. Image oracle is results with predicted object identity based on ground truth image-level annotations, identity oracle is results with  ground truth object identities based on predicted image-level instances.}
\label{tab:oracle}
\begin{tabular}{|l|l|l|}
\hline
                & AP   & AR\textsubscript{10} \\ \hline
Image Oracle    & 78.7 & 83.7    \\ \hline
Identity Oracle & 31.5 & 34.6    \\ \hline
\end{tabular}
\end{table}

\section{Conclusions}\label{sec:conclude}
We present a new task named video instance segmentation and an accompany dataset named YouTubeVIS in this work. The new tasks is a combination of object detection, segmentation, and tracking, which poses specific challenges given the rich and complex scenes. We also propose a new method combining single-frame instance segmentation and object tracking, which aims to provide some early explorations towards this task. There are a few interesting future directions: object proposal and detection with spatial-temporal features, end-to-end trainable matching criterion, and incorporating motion information for better recognition and identity association.
We believe the new task and new algorithm will innovate the research community on new research ideas and directions for video understanding. 

{\small

\bibliographystyle{ieee_fullname}
\bibliography{egbib}
}

\end{document}